\pdfoutput=1
\documentclass[runningheads]{llncs}
\usepackage{graphicx}

\usepackage{tikz}
\usepackage{comment}
\usepackage{amsmath,amssymb} %
\usepackage{color}
\usepackage[acronym]{glossaries}
\usepackage{xspace}
\usepackage{todonotes}
\usepackage{array}
\usepackage{microtype}
\usepackage{subcaption}
\usepackage{booktabs, multirow} %
\usepackage{soul}%
\usepackage{changepage,threeparttable} %
\usepackage{enumitem}
\usepackage{relsize}

\newacronym{nerf}{NeRF}{neural radiance field}
\newacronym{hmd}{HMD}{head-mounted display}
\newacronym{mlp}{MLP}{multilayer perceptron}
\newacronym{nsr}{NSR}{neural scene representation}
\newacronym{vr}{VR}{virtual reality}
\newacronym{ar}{AR}{augmented reality}
\newacronym{ibr}{IBR}{image-based rendering}
\newacronym{mpi}{MPI}{multi-plane image}
\newacronym{nsvf}{NSVF}{neural sparse voxel field}
\newacronym{srn}{SRN}{scene representation network}
\newacronym{nv}{NV}{neural volume}
\newacronym{derf}{DeRF}{decomposed radiance field}
\newacronym{llff}{LLFF}{Local Light Field Fusion}
\newacronym{nex}{NeX}{Neural Basis Expansion}
\newacronym{donerf}{DONeRF}{depth oracle neural radiance field}
\newacronym{ours}{AdaNeRF}{adaptive sampling for neu}
\newacronym{ingp}{INGP}{instant neural graphics primitives}

\newcommand{\ours}{AdaNeRF\xspace}
\newcommand{\donerf}{DONeRF\xspace}
\newcommand{\donerftermi}{TermiNeRF*\xspace}

\setglossarysection{paragraph}
\makeglossaries

\makeatletter
\newcommand{\printfnsymbol}[1]{%
        \textsuperscript{\@fnsymbol{#1}}%
}
\makeatother

\begin{document}

\newcommand{\myitem}[2]{\noindent \textbf{({#1})\ #2:}}
\newcommand{\zhaoyang}[1]{\textcolor{red}{Zhaoyang:{#1}}}
\newcommand{\markus}[1]{\textcolor{blue}{Markus:{#1}}}
\newcommand{\thomas}[1]{\textcolor{green}{Thomas:{#1}}}
\newcommand{\revised}[1]{{{#1}}}
\newcommand{\MZ}[1]{\textcolor{red}{MZ:{#1}}}

\pagestyle{headings}
\mainmatter
\def\ECCVSubNumber{6513}  %

\title{AdaNeRF: Adaptive Sampling for Real-time Rendering of Neural Radiance Fields}

\titlerunning{AdaNeRF}
\author{Andreas Kurz\inst{1}\thanks{Authors contributed equally to this work.} \and
Thomas Neff\inst{1}\printfnsymbol{1} \and
Zhaoyang Lv\inst{2} \and \\
Michael Zollhöfer\inst{2} \and 
Markus Steinberger\inst{1}
}
\authorrunning{Kurz et al.}
\institute{Graz University of Technology, Austria \and
Reality Labs Research, USA}

\maketitle

\vspace{-0.05cm}
\begin{changemargin}{-0.05cm}{-0.05cm} 
\begin{abstract}
Novel view synthesis has recently been revolutionized by learning neural radiance fields directly from sparse observations.
However, rendering images with this new paradigm is slow due to the fact that an accurate quadrature of the volume rendering equation requires a large number of samples for each ray.
Previous work has mainly focused on speeding up the network evaluations that are associated with each sample point, e.g., via caching of radiance values into explicit spatial data structures, but this comes at the expense of model compactness.
In this paper, we propose a novel dual-network architecture that takes an orthogonal direction by learning how to best reduce the number of required sample points.
To this end, we split our network into a sampling and shading network that are jointly trained.
Our training scheme employs fixed sample positions along each ray, and incrementally introduces sparsity throughout training to achieve high quality even at low sample counts.
After fine-tuning with the target number of samples, the resulting compact neural representation can be rendered in real-time.
Our experiments demonstrate that our approach outperforms concurrent compact neural representations in terms of quality and frame rate and performs on par with highly efficient hybrid representations.
Code and supplementary material is available at \small{\url{https://thomasneff.github.io/adanerf}}.
\end{abstract}
\end{changemargin}

\begin{figure}[t]
	\centering
	\includegraphics[width=0.97\linewidth]{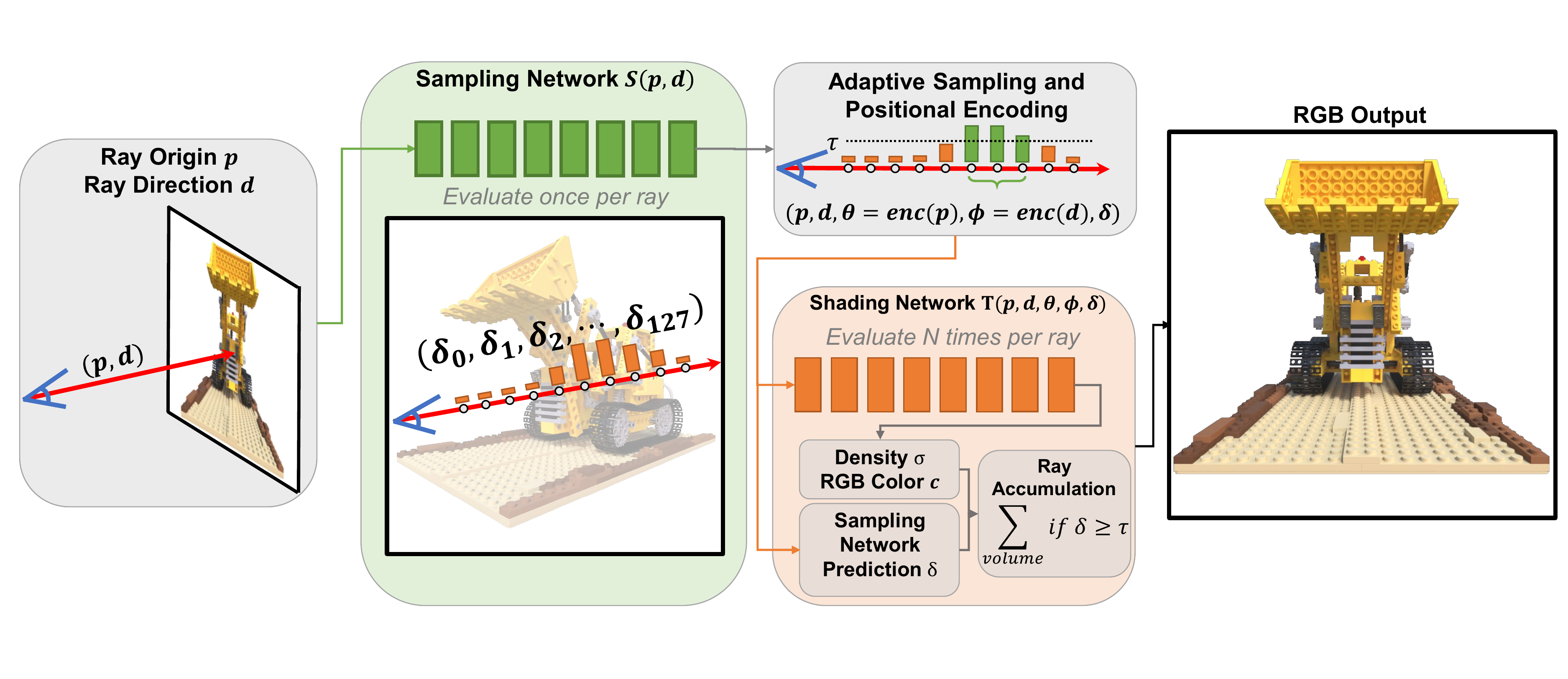}
	\caption{\ours employs a single-evaluation \emph{sampling network} and a multi-evaluation \emph{shading network} to significantly reduce the number of required network evaluations per view ray. For each ray, the sampling network predicts a vector of estimated sample densities $\boldsymbol{\delta}$ that correspond to exactly one sample location each. We threshold the predicted $\boldsymbol{\delta}$ to cull away samples that are expected to have only minor contribution and only proceed to evaluate the shading network for this small subset of samples along the ray. Finally, when accumulating the outputs of the shading network along the ray, we additionally multiply the density $\boldsymbol{\sigma}$ predicted by the shading network with the density $\boldsymbol{\delta}$ predicted by the sampling network. This enables gradients of the RGB output loss to flow back to the sampling network, enabling end-to-end training of the full pipeline.}
	\label{fig:method_main}
\end{figure}

\section{Introduction}
\label{sec:introduction}
The introduction of neural radiance fields~\cite{mildenhall2020nerf} pushed the boundaries of modern computer graphics and vision by improving the state of the art of applications such as 3D reconstruction~\cite{takikawa2021nglod}, rendering~\cite{yu2021plenoxels,yu2021plenoctrees,mueller2022instant,neff2021donerf}, animation~\cite{park2021nerfies,pumarola2020d} and scene relighting~\cite{martinbrualla2020nerfw}.
Furthermore, since the introduction of neural radiance fields, a significant amount of research focused on improving the resulting image quality~\cite{barron2021mipnerf}, training speed~\cite{lindell2020autoint,mueller2022instant}, and inference performance~\cite{neff2021donerf,mueller2022instant,piala2021terminerf,Garbin_2021_ICCV,reiser2021kilonerf,hedman2021snerg,yu2021plenoctrees}.
Thus, real-time rendering of photorealistic neural radiance fields is now possible on standard consumer GPUs.

However, current real-time renderable neural radiance fields either require large amounts of memory, a restricted training data distribution or bounded training data, and it is unclear how to find an efficient compromise between those limitations.
Explicit data structures have difficulties accounting for unbounded scenes, and storing large amounts of \glspl{nerf} in sparse grids~\cite{hedman2021snerg,Garbin_2021_ICCV}, trees~\cite{yu2021plenoctrees} or hash tables~\cite{mueller2022instant} consumes prohibitive amounts of memory if multiple neural radiance fields need to be accessed in quick succession, such as in a streaming scenario. 
At a compact memory footprint, previous work relied on reducing the number of samples per view ray via dedicated \emph{sampling networks} that estimate suitable sample locations along each view ray to improve rendering speed~\cite{piala2021terminerf,neff2021donerf,lindell2020autoint}.
These sampling networks are commonly trained via supervision from depth~\cite{neff2021donerf} or the predicted density of a neural radiance field~\cite{piala2021terminerf}, thus requiring additional time-consuming preprocessing or pretraining steps.
Alternatively, sampling networks can also be used to learn segments along each ray using an integral network ~\cite{lindell2020autoint}.
Although this improves efficiency at a slight loss in quality, constructing these integral networks drastically increases the complexity and duration of training.
Finally, light field networks~\cite{sitzmann2021lfns} only evaluate a single sample per ray by parameterizing the input ray using Plücker coordinates,
but learning such a light field typically requires meta-learning to achieve sufficient quality even on small toy-datasets.

In this paper, we introduce \ours, a compact dual-network neural representation that is optimized end-to-end, and fine-tuned to the desired performance for real-time rendering.
The first \emph{sampling network} predicts suitable sample locations using a single evaluation per view ray, while the second \emph{shading network} adaptively shades only the most significant samples per ray.
In contrast to previous methods based on sampling networks, \ours does not require any preprocessing, pretraining or special input parametrizations, lowering the overall complexity.
We use fixed, discrete sample locations along each ray to set up a soft student-teacher regularization scheme by multiplying the predicted density of our sampling network with the output density of the shading network.
Thus, both networks can modify the final RGB output and gradients flow throughout the whole pipeline.
We ensure sparsity within our sampling network via a \emph{4-phase training scheme}, after which we fine-tune our shading network to the desired sample counts for real-time rendering.
We \emph{adaptively sample} our shading network for each individual ray---we only evaluate the shading network for the most important samples, as predicted by the sampling network.
The resulting sparse, dual-network pipeline can be rendered in real-time on consumer GPUs using our custom real-time renderer based on CUDA and TensorRT.

Our experimental results demonstrate the benefits of \ours compared to prior arts on a variety of datasets, including large, unbounded scenes.
First, the adaptive sampling in \ours significantly increases the sampling efficiency of raymarching-based neural representations. 
Second, \ours outperforms previous sampling network based approaches in both rendering speed and quality with the same compact memory footprint.
Finally, we qualitatively show that multiple AdaNeRFs can scale to complex scenes of arbitrary size. %

In summary, we make the following contributions:
\begin{itemize}[noitemsep, topsep=0pt]
\item
A novel dual-network architecture to jointly learn sampling and shading networks for compact real-time neural radiance fields, outperforming existing sampling-network based approaches.
\item
An additional adjustable adaptive sampling scheme to only shade the most significant samples per ray, further improving quality and efficiency at identical average sample counts.
\item
A real-time rendering implementation that relies on dynamic, sparse sampling of our compact dual-network representation, targeting a sweet spot in the trade-off between performance, quality, and memory.
\end{itemize}

\section{Related Work}
\label{sec:related_work}
Since the introduction of \gls{nerf}~\cite{mildenhall2020nerf}, coordinate-based neural radiance fields have improved the state-of-the-art across many domains, including dynamic scene modeling \cite{Gao-ICCV-DynNeRF,du2021nerflow,park2021nerfies,tretschk2021nonrigid,li2021neural,Xian2020stnerf,Pumarola2020DNeRF,Park2021hypernerf,Zhang2021stnerf}, animatable avatars and scenes \cite{lombardi2021mixture,Peng2021animatable,Chen2021animatable,Liu2021neuralactor,Yang2021objectnerf}, relightable objects \cite{Srinivasan20arxiv_NeRV,Boss20arxiv_NeRD}, and object reconstruction \cite{Oechsle2021unisurf,Xie2021fignerf,Wang2021neus,Yariv2021volume}. 
The quality of object captures can be improved by incorporating different scales into the encoding, reducing aliasing and sampling artifacts when novel views are generated~\cite{barron2021mipnerf}.
While this is mostly restricted to single-object captures, recent research has also investigated the reconstruction from unconstrained  images~\cite{martinbrualla2020nerfw} and large, unbounded scenes~\cite{zhang2020nerf,barron2021mipnerf360}. However, most advances of neural radiance fields focus on improving the output quality, with many of these advancements being infeasible to compute in real-time on consumer GPUs. 

Our work is closely related to the advancement of neural radiance fields towards real-time rendering performance, which can be categorized into three domains: (1) Decomposed neural radiance fields. (2) Baking, caching or precomputing weights into an explicit spatial data structure, and (3) improving the sampling efficiency and reducing the total number of samples that are computed per frame.

\paragraph{Decomposed neural radiance fields.} 
By splitting a single MLP into many separate MLPs~\cite{reiser2021kilonerf,rebain2020derf,fang2021neusample,Rebain20arxiv_derf}, subdivided scene grids \cite{liu2020neural} or primitives \cite{lombardi2021mixture}, both the quality and efficiency of rendering can be increased. 
Such a composition of scenes can represent scenes at city-scale~\cite{turki2021meganerf,tancik2022blocknerf}. 
Although these representations are useful to render single objects \cite{reiser2021kilonerf} or human avatars \cite{lombardi2021mixture} in real-time, a real-time, scene-level representation has yet to be demonstrated.

\paragraph{Baking radiance fields.} The radiance field can be stored inside a 3D grid~\cite{Garbin_2021_ICCV}, inside a sparse voxel octree~\cite{yu2021plenoctrees,liu2020neural} that does not even require any neural networks~\cite{yu2021plenoxels}, or inside a sparse grid~\cite{hedman2021snerg}.
These methods run in real-time at a significantly increased memory footprint, which can be prohibitively expensive for scenarios such as streaming, where real-time swapping of neural radiance fields is desired. 
The concurrently introduced Instant-NGP~\cite{mueller2022instant} combines a hierarchical spatial hash table encoding with small MLPs to learn and render a scene representation in real-time.
However, it is still unclear how it performs in demanding real-time scenarios and how to optimally tune its hash table size to still be as compact as fully neural representations.

\paragraph{Improving the sampling efficiency of neural radiance fields.}
Recent work drastically improved the sampling efficiency of \gls{nerf} while keeping the same compact memory footprint.
DONeRF~\cite{neff2021donerf} proposed a reduction in overall sample count by swapping the coarse network of the original \gls{nerf} with a \emph{depth oracle network}, which can provide suitable sample locations for the second \emph{shading network}, thus reducing the number of samples per ray by up to $128\times$. 
However,
it cannot be trained end-to-end, and struggles without reliable depth information.
Similarly to DONeRF, TermiNeRF~\cite{piala2021terminerf} uses a \emph{sampling network} that is conditioned on the density of a pre-trained \gls{nerf}.
TermiNeRF uses the whole range of samples without the need for a depth map, which can improve quality in geometrically ambiguous scenarios.
Although it can technically be trained end-to-end, \mbox{TermiNeRF} requires a suitable pretrained NeRF for initialization of its color network to achieve the best results.
AutoInt~\cite{lindell2020autoint} approaches sampling networks differently by predicting the lengths of segments along each ray.
By predicting ray segments instead of samples, subsequent \emph{integral} networks can efficiently predict the density and color of each ray segment, reducing the number of network evaluations.
However, the training procedure is significantly longer and more complex.
Light field networks~\cite{sitzmann2021lfns,Attal2021} reduce the number of network evaluations to a single sample per ray by directly mapping a view ray to the observed color. 
Although such an approach is advantageous in terms of memory footprint and rendering efficiency, without a meta-learned multi-view consistency prior, it fails to synthesize novel views in real-world scenes, and is thus not suitable for real-time novel view synthesis tasks for real world captures.

With \ours, we follow up on DONeRF \cite{neff2021donerf}, TermiNeRF \cite{piala2021terminerf}, and AutoInt \cite{lindell2020autoint}, and demonstrate that our approach is end-to-end trainable and more robust across a variety of training setups. 
Our approach achieves higher quality with faster real-time rendering performance by drastically reducing the amount of required samples, while at the same time keeping the representation compact.
\section{Method}
\label{sec:method}

\ours consists of a fully end-to-end trainable pipeline that can be rendered in real-time.
We replace the coarse network of the original NeRF by a \emph{sampling network} $S$ that is only evaluated once per ray, minimizing the number of network evaluations to generate the final image.
The sampling network takes the ray origin $\mathbf{p}$ and the ray direction $\mathbf{d}$ as input.
The output of the sampling network is a vector of predictions $\boldsymbol{\delta}$, corresponding to the predicted importance of samples along each ray.
The \emph{shading network} $T$ takes the prediction of the sampling network, positionally encodes the samples with the largest contribution and outputs their density $\boldsymbol{\sigma}$ and color $\mathbf{c}$.
By evaluating the sampling network once, a majority of samples with low contributions can be culled, increasing the overall efficiency of the pipeline.
Figure~\ref{fig:method_main} shows an overview of our dual-network setup.

\subsection{End-to-end Trainable Sampling Network}

We propose to multiply the predicted per-sample density $\delta_i$ of the sampling network with the predicted per-sample density $\sigma_i$ of the shading network. 
This formulation allows backpropagation to reach the sampling network.
This is possible by using fixed sample locations along each ray, and placing exactly one sample in the center of each cell when discretizing the space along each ray. 
In contrast to previous work based on sampling networks, we do not require ground truth depth~\cite{neff2021donerf}, and we avoid distinctly separated training steps~\cite{neff2021donerf,piala2021terminerf}.

We modify the standard ray accumulation function~\cite{mildenhall2020nerf} to include an additional multiplication by the outputs $\delta_i$ of the sampling network
\revised{%
\begin{equation}
	\label{eqn:density_mult}
	\hat{C}(\mathbf{r}) = \sum_{i=1}^{N} T_i (1 - \exp(-\delta_i \sigma_i t_i)) \mathbf{c}_i, \; \text{where} \; T_i = \exp \left( - \sum_{j=1}^{i - 1} \delta_j  \sigma_j t_j \right),
\end{equation}
}
where $\hat{C}$ is the estimated, accumulated color, $N$ is the number of samples along the ray, $T_i$ is the accumulated transmittance along the ray, $t_i$ is the distance between adjacent samples and $\sigma_i$ is the output density of the shading network for sample $i$. 
The introduction of the multiplication by $\delta_i$ enables the sampling network to directly increase or decrease the importance of samples via its prediction, and to receive gradients from the MSE color reconstruction loss.

\subsection{Sparse Adaptive Sampling Network Distillation}
\label{sec:training_scheme}
The modification of the ray accumulation alone does not ensure that the sampling network outputs sparse predictions---it might just as well always output the $\mathbf{1}$ vector, leading the shading network to place one sample in each cell, effectively ignoring the sampling network prediction.
We disentangle $\boldsymbol{\delta}$ from the shading network density by introducing \emph{sparsity} into the sampling network, which forces the network to select only the most important density values.

\begin{figure}[t]
	\centering
	\includegraphics[width=1.0\linewidth]{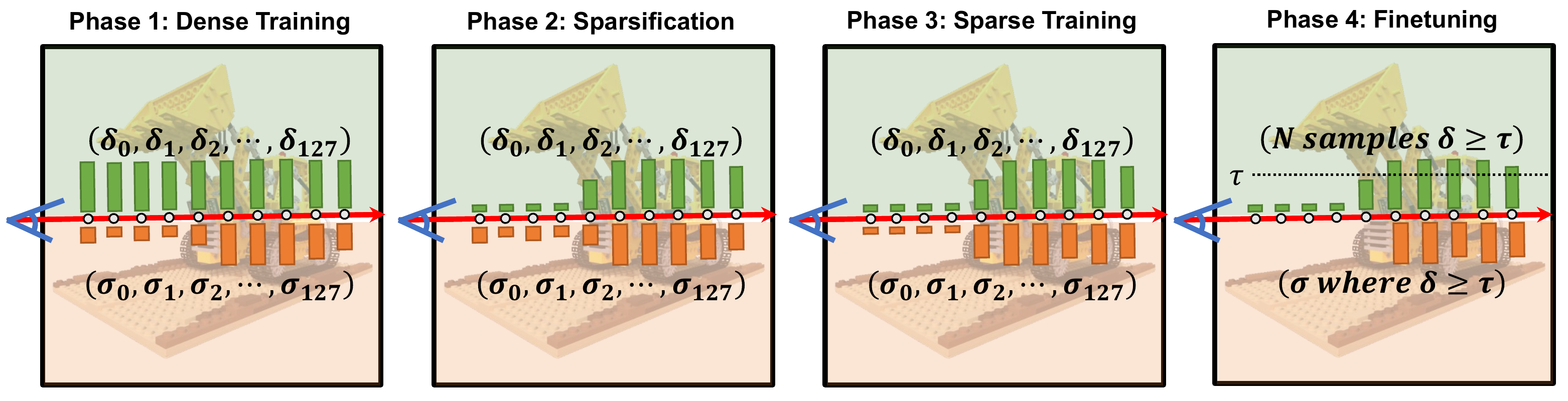}
	\caption{The 4-phase training scheme of \ours. First, \emph{dense training} forces the sampling network predictions close to $\boldsymbol{\delta} = \mathbf{1}$, enabling the shading network to densely sample the underlying scene. The \emph{sparsification} phase then forces a majority of the sampling network predictions towards $\boldsymbol{\delta} = \mathbf{0}$, leaving only the most significant samples. The \emph{sparse training} phase then adjusts the shading network to the newly sparsified sampling network. Finally, in the \emph{finetuning} phase, we adaptively place samples for all sampling network predictions $\boldsymbol{\delta} \geq \tau$ up to a desired maximum number to further optimize for real-time rendering.}
	\label{fig:training_scheme}
\end{figure}

\ours trains the sampling network and the shading network end-to-end and progressively reduces the required samples per view ray.
The shading network is trained via a standard MSE loss on the accumulated RGB color.
The sampling network loss is composed of a sparsity loss which includes an $\ell_1$-loss that matches the output density of the shading network and an additional \emph{density multiplication term} derived from the MSE loss of the shading network:
\begin{equation}
	\label{eqn:sampling_network_loss}
	l_{sampling}(\boldsymbol{\delta}, \boldsymbol{\sigma}, \mathbf{c}) = \lambda_{0} \cdot l_{mse}(\boldsymbol{\delta}, \boldsymbol{\sigma}, \mathbf{c}) + \lambda_{1} \cdot l_{sparsity}(\boldsymbol{\delta}, \boldsymbol{\sigma}).
\end{equation}

\noindent \ours uses a soft student-teacher regularization training scheme with $4$ phases. We illustrate the training scheme in Fig.~\ref{fig:training_scheme} and describe each phase in detail:

\paragraph{Dense Training.}
\label{sec:dense_training} 
The initial dense training phase establishes the \emph{teacher}, by encouraging the sampling network to output dense predictions via an $\ell_1$-loss of all its outputs towards $\mathbf{1}$.
\revised{
In practice, this phase could be replaced by initializing the network weights to output $\mathbf{1}$ directly to increase training speed at a potential loss of propagated information to the sampling network.
In either way, the shading network samples the full input space to provide an initial estimate of the scene, which prevents both networks from collapsing in the later stages.}

\paragraph{Sparsification.}
\label{sec:sparsification} 
The second phase introduces an additional $\ell_1$-loss to the sampling network that forces the majority of predictions towards $\mathbf{0}$. 
We linearly blend between forcing the sampling network outputs towards $\mathbf{1}$ and $\mathbf{0}$ over the course of this phase, and additionally blend in a soft student-teacher regularization loss via an $\ell_1$-term that encourages the sampling network outputs $\boldsymbol{\delta}$ to follow a similar distribution as $\boldsymbol{\sigma}$.
From iteration $t_0$ over the duration of $t_d$ iterations (and the current iteration given as $t_c$), we define the sparsification loss as
\begin{equation}
	\label{eqn:sparsification_loss}
	\begin{aligned}
	l_{sparsity}(\boldsymbol{\delta}, \boldsymbol{\sigma}) & = \lambda \cdot \frac{1}{N} \left( \sum_{i=1}^{N}   |\delta_i - 0| +  |\sigma_i - \delta_i|  \right) + (1 - \lambda) \cdot \frac{1}{N} \sum_{i=1}^{N} |\delta_i - 1|,\\  & \text{where} \; \lambda = \frac{t_c - t_0}{t_d}.
	\end{aligned}
\end{equation}
The $\ell_1$-term $|\sigma_i - \delta_i|$ ensures that the sampling network does not collapse to a single constant $\mathbf{0}$ or $\mathbf{1}$ vector, forcing the sampling network to follow the established scene representation of the shading network. %
The sparsification phase gradually increases the sparsity of the sampling network resulting in fewer significant outputs (which subsequently have zero contribution during ray accumulation).

\paragraph{Sparse Training.}
\label{sec:sparse_training}
To allow the shading network to take advantage of the sparsification of the sampling network, we lock the sampling network's weights during sparse training.
Although the shading network is still queried for all samples along each ray (as in the dense training phase) it is now free to alter the output for samples that are already dampened by the sampling network (due to the density multiplication).
This enables the network to focus its capacity on those samples that actually contribute to the output.

\paragraph{Fine-tuning.}
\label{sec:fine_tuning} 
\revised{We fine-tune the shading network for a desired maximum number of samples per ray; typically 2, 4, 8 or 16.
This phase is fast, as the number of samples per ray is small.
Fine-tuning can increase quality as it completely removes samples that hardly contribute to the final output and allows the shading network to focus on the contributing samples only.  
Note that this phase results in separate shading networks for each maximum sample count, while all rely on the same sampling network.
}

\paragraph{Real-time Rendering with Adaptive Sampling.}
\label{sec:adaptive_sampling} 
We can further improve performance by enabling variable sample counts per ray.
This adaptive sampling scheme exploits the fact that \ours uses fixed sample locations along the ray that can at most contain exactly one sample.
First, we add an adaptive sampling threshold $\tau$ that defines the cutoff point for the sampling network's predictions $\boldsymbol{\delta}$.
This enables us to save shading network evaluations in regions that do not require more than a few samples (such as a uniformly colored sky or simple geometric objects), which in turn increases the overall efficiency of our pipeline.
Then, we limit the maximum number of allowed samples to $N_{max}$, and distinguish between the following cases, depending on the number of sampling network predictions $N_s$ that exceed the threshold $\tau$:
\begin{enumerate}
	\item $N_s = 0:$ If no sampling network predictions $\delta_i$ exceed $\tau$, we place one sample at the center of the ray segment corresponding to the sampling network's largest prediction.
	\item $N_s \leq N_{max}:$ If the number of sampling network predictions $\delta_i$ that exceed $\tau$ is at most $N_{max}$, we place samples at the center of all of their segments.
	\item $N_s > N_{max}:$ If the number of sampling network predictions $\delta_i$ that exceed $\tau$ is more than our maximum number of allowed samples $N_{max}$, we place one sample each at the center of the $N_{max}$ largest predictions.
\end{enumerate}
This adaptive sampling scheme can be efficiently implemented on GPUs using warp communication primitives, enabling further efficiency gains compared to typical importance sampling setups that first need to generate a cumulative distribution function from a probability density function.
Note that our approach is the first neural representation that relies on volume integration and (1) can go down to a single sample per ray and (2) supports variable sample counts per ray without the need for a spatial data structure.

\section{Evaluation}
\label{sec:evaluation}

\paragraph{Implementation Details.} 
We follow the network architecture of \donerf~\cite{neff2021donerf}, using MLPs consisting of $8$ layers with $256$ units for both the sampling and shading networks. 
For the \donerf dataset, we logarithmically space samples along each ray and unify rays~\cite{neff2021donerf}. %
For the LLFF dataset, we sample in normalized device coordinates~\cite{mildenhall2020nerf}.
We use Adam~\cite{kingma2015adam} with a learning rate of $5e^{-4}$ in training.
We configure our 4-phase training scheme (Section~\ref{sec:training_scheme}) in the following order:
 $25$k iterations of \emph{dense training},  $50$k iterations \emph{sparsification}, $225$k iterations \emph{sparse training}, and $300$k iterations \emph{fine-tuning}.
We vary the adaptive sampling threshold $\tau$ in comparison to other baselines at similar average sample counts. 
\revised{As a starting point,} for the MSE loss of the sampling network (Equ.~\ref{eqn:density_mult}, Equ.~\ref{eqn:sampling_network_loss}) we use $\lambda_{0} = 0.001$, and for the sparsity loss of the sampling network (Equ.~\ref{eqn:sparsification_loss}, Equ.~\ref{eqn:sampling_network_loss}) we use $\lambda_{1} = 1.0$. 
\revised{Please refer to the supplementary material for per-scene loss weights that were found by grid search and used in our evaluation.}
Finally, for the real-time performance comparison, we implemented a custom real-time renderer using CUDA and TensorRT to take advantage of our adaptive sampling strategy.
All results were evaluated on a single Nvidia RTX 3090.

\subsection{Ablation Studies} 
\label{sec:e2e_ablations}
We provide an ablation study to validate the design of our 4-phase training scheme~(Section~\ref{sec:training_scheme}) and our adaptive sampling strategy~(Section~\ref{sec:adaptive_sampling}), averaged across the \emph{Pavillon} scene of  \donerf dataset. 
\revised{The number of iterations for each phase was determined in small-scale experiments and could be further optimized for training speed or image quality.}

\paragraph{Training Scheme}
Tab.~\ref{tbl:ablation_training} shows the ablation of our training scheme.
Without (2) dense training or (3) sparsification, we observe a minor degradation in quality. %
If dense training is skipped, the shading network provides less accurate information to the sampling network; if sparsification is skipped, the sampling network is abruptly forced to be sparse \revised{by switching from ``fully dense'' to ``fully sparse'' training immediately instead of blending between them}, losing potentially important samples in the process.
Removing the (4) density multiplication in the ray accumulation function (Equ.~\ref{eqn:density_mult}) results in the sampling network collapsing to a constant output---the $\ell_1$-loss as the only supervision signal is insufficient to stabilize the sampling network. 
Similarly, using (5) $\ell_1$-loss supervision from the shading network as the sole optimization criterion (Equ.~\ref{eqn:sparsification_loss}) leads to the sampling network collapsing towards the mean density of all rays.
Removing (6) the shading density supervision $\ell_1$-term from Equ.~\ref{eqn:sparsification_loss} still produces reasonable sampling networks, at a quality degradation due to the lack of additional supervision.
\revised{Finally, removing (7) the sparse training directly fine-tunes after sparsification.
The resulting shading networks are not adapted to the sparsified sampling networks, significantly reducing quality. }

\paragraph{Adaptive Sampling}
We sweep the threshold $\tau$ between $[0.05, 0.40]$
and compare the resulting quality against fixed sample counts of $N = [8, 16]$, see Tab.~\ref{tbl:ablation_adaptive_sampling}.
Compared to the fixed sample count of $N = 8$, the adaptive variant reaches similar quality between $5.07$ and $6.16$ samples per pixel, showing the increased efficiency even at lower sample counts. 
As average sample counts increase, the sampling network has much more freedom in placing the samples, and thus can outperform the quality of $N = 16$ fixed samples at just $7.76$ samples. 

\begin{table}[t]\centering
	\caption{Ablation of the 4-phase training scheme of \ours, using two maximum sample counts of $N_{max}=[2, 4]$ on the \emph{Pavillon} scene of the \donerf dataset. 
	}\label{tbl:ablation_training}
	\scalebox{0.9}{
	\begin{tabular}{l|cc|cc}\toprule
		& \multicolumn{2}{c|}{$N_{max} = 2$} & \multicolumn{2}{c}{$N_{max} = 4$} \\
		Method &$N$ / Ray &PSNR $\uparrow$ &$N$ / Ray &PSNR $\uparrow$ \\\midrule
		1) \ours &2.00 &\textbf{28.25} &3.99 &\textbf{29.33} \\\midrule
		2) No dense training &1.99 &27.69 &3.54 &29.23 \\
		3) No sparsification blending &1.66 &27.33 &2.75 &28.44 \\
		4) No weight multiplication &1.00 &24.75 &1.00 &24.73 \\
		5) Only shading density supervision &1.00 &24.72 &1.00 &24.73 \\
		6) No shading density supervision &2.00 &27.82 &4.00 &28.86 \\
		7) No sparse training &2.00 &21.38 &4.00 &28.39 \\
		\bottomrule
	\end{tabular}
	}
\end{table}
\begin{table}[t]\centering
	\caption{
	On the \emph{Pavillon} scene of the \donerf dataset, our adaptive sampling scheme manages to achieve higher quality at average sample counts of $6.16$ and $7.76$, compared to fixed  sample counts of $N = [8, 16]$. }\label{tbl:ablation_adaptive_sampling}
	\scalebox{0.9}{
	\begin{tabular}{l|rr|rrrrrrrr}\toprule
		&\multicolumn{2}{c|}{Fixed} &\multicolumn{8}{c}{Adaptive} \\\midrule
		Threshold $\tau$ &- &- &0.05 &0.1 &0.15 &0.2 &0.25 &0.3 &0.35 &0.4 \\
		Samples per Ray &16.00 &8.00 &12.10 &11.89 &11.63 &11.03 &9.73 &7.76 &6.16 &5.07 \\
		PSNR$\uparrow$ &30.89 &30.40 &31.66 &31.66 &31.68 &31.66 &31.62 &31.07 &30.64 &30.24 \\
		\bottomrule
	\end{tabular}
	}
\end{table}

\subsection{Results}

We show a quantitative and qualitative evaluation of \ours on a variety of datasets against several baseline methods.
We measure the quality of the rendered images in PSNR, and report the number of parameters required to store each method (using uncompressed 32-bit floating point) to evaluate compactness.
We further present real-time rendering timings measured via TensorRT and CUDA. Please refer to the supplementary material for more visual comparisons, \revised{a discussion on training speed and a discussion on how to interleave multiple \ours for larger scenes.}

\paragraph{Datasets.} 
We evaluate our method using the following datasets.
\begin{itemize}[noitemsep, topsep=0pt]
    \item The \textbf{\donerf~\cite{neff2021donerf} dataset} contains synthetic indoor and outdoor scenes of small to very large scales that are path-traced using Blender at a resolution of $800\times800$, with the cameras aimed at the forward hemisphere of their bounding box. %
    \item The \textbf{LLFF~\cite{mildenhall2019llff} dataset} contains forward-facing real-world scenes captured using a handheld camera, which we scale to a resolution of $1008\times756$. We follow the convention~\cite{mildenhall2019llff} of holding out every 8th image for testing.
\end{itemize}

\paragraph{Baselines.} 
Besides comparing to NeRF \cite{mildenhall2020nerf}, we compare \ours to related work that focused on improving sampling efficiency and rendering performance:
\begin{itemize}[noitemsep, topsep=0pt]
    \item \textbf{\donerf}~\cite{neff2021donerf}: \donerf uses a depth oracle network trained on depth maps to improve sampling efficiency. For all experiments, we train the oracle network using depth maps extracted from a pre-trained coarse NeRF.
    \item \textbf{\donerftermi}~\cite{piala2021terminerf}: \donerftermi learns a sampling network based on the density of a pre-trained NeRF. We follow the input encoding of \donerf, and further use $128$ fixed sample locations for the targets extracted from the pre-trained coarse NeRF, avoiding resampling and filtering of the targets. 
    \item \textbf{AutoInt}~\cite{lindell2020autoint}: AutoInt learns automatic integration via a sampling network. We compare \ours to AutoInt on a lower resolution version of the LLFF dataset, which was provided in the authors' original paper. 
    \item \textbf{Plenoxels}~\cite{yu2021plenoxels}: Plenoxels uses a sparse grid with trilinear interpolation to directly learn a scene representation via spherical harmonics, without neural networks. For unbounded scenes, Plenoxels uses a multi-sphere-image background model in combination with its sparse foreground grid model.
    \revised{%
    \item \textbf{Instant-NGP}~\cite{mueller2022instant}: Instant-NGP uses a hierarchical hash table to store most of its representation, with only tiny MLPs used to trilinearly interpolate the hash table entries along each ray. We show results for the default hash table size of $2^{19}$, as well as the authors' suggested smaller and faster alternative of $2^{14}$.}
\end{itemize}
For both \donerf and \donerftermi, we first train a coarse-only NeRF at $128$ samples per ray with $8$ hidden layers with $256$ units each. We compare \ours to \donerf and \donerftermi with fixed sample counts of $N = [2, 4, 8, 16]$.
For Plenoxels we use configurations provided by the authors: \emph{Plenoxels} uses a sparse grid resolution of $256^3$. \emph{Plenoxels-MSI} adds a background model with $64$ layers. %
\emph{Plenoxels-Large} uses the authors' provided checkpoints for the LLFF dataset, which are significantly more dense.
For all baselines, we use the available open source code, and the authors' suggested settings unless otherwise specified.

\subsubsection{Quality}
\label{sec:quality}
We present average output quality, memory footprint and render times for the \donerf and LLFF datasets in Table~\ref{tbl:eval_results}, and
example outputs in Figure~\ref{fig:eval_qualitative}. \revised{Additional examples, per-scene data, depth reconstructions and sample placement visualizations can be found in the supplementary material.}

\revised{
For the \donerf dataset, Instant-NGP-$2^{19}$ achieves the best quality, followed by all NeRF-based approaches with similar quality.
Considering run-time, \ours shows the best tradeoff, allowing to choose between very fast rendering (at $3.7$ samples) and competitive quality or high-quality (at $7.0$ samples) and $2\times$ speed improvement over \donerf and \donerftermi at the same quality.
\ours only falls behind \donerf and \donerftermi in image quality at extremely low sample counts while achieving greater speed improvement, suggesting that \ours operates most variably at a slightly higher sample counts. 
Considering memory foot footprint, \ours achieves equal or better quality than Plenoxels at a $48-215\times$ reduction in memory and similar run-time.
Instant-NGP-$2^{19}$ is similar to \ours considering all three tradeoffs: it achieves higher quality at a higher memory and run-time cost, or similar quality with lower memory but higher run-time.
For the LLFF dataset, the highest quality is achieved by NeRF, Plenoxels-Large, and \ours at $10.2$ samples.
Compared to other sampling-network based approaches, \ours clearly outperforms the state-of-the-art, achieving better quality than \donerftermi at less than half the sample count and frame time. 
Again, considering memory footprint and performance, both NeRF and Plenoxels show significant drawbacks compared to \ours: Plenoxels requires $3.6$ GB of memory for its representation and NeRF takes $2.8$ seconds to render a single frame. 
Interestingly, Instant-NGP ($2^{19}$ and $2^{14}$) perform worse than \ours  for this data set.
Compared to AutoInt, \ours achieves better quality at much faster rendering speeds.
In summary, our adaptive fully neural representation shows state-of-the-art image quality at equal or better run-time performance and memory footprint, without requiring explicit data structures.
}

\newcommand{\reducedstrut}{\vrule width 0pt height 1.0\ht\strutbox depth 1.0\dp\strutbox\relax}
\newcommand{\gold}[1]{%
	\begingroup
	\setlength{\fboxsep}{1pt}%
	\colorbox[HTML]{AF9500}{\reducedstrut\textcolor[HTML]{FFFFFF}{\textbf{#1}}\/}%
	\endgroup
}
\newcommand{\silv}[1]{%
	\begingroup
	\setlength{\fboxsep}{1pt}%
	\colorbox[HTML]{A4A4A4}{\reducedstrut\textcolor[HTML]{FFFFFF}{\textbf{#1}}\/}%
	\endgroup
}
\newcommand{\bron}[1]{%
	\begingroup
	\setlength{\fboxsep}{1pt}%
	\colorbox[HTML]{6A3805}{\reducedstrut\textcolor[HTML]{FFFFFF}{\textbf{#1}}\/}%
	\endgroup
}

\begin{table}[t]\centering
	\caption{Image quality, render time and memory footprint comparison on the \donerf~\cite{neff2021donerf} and LLFF~\cite{mildenhall2019llff} datasets. Best results are displayed as \gold{Top 1}, \silv{Top 2} and \bron{Top 3} per category.}\label{tbl:eval_results}
	\scriptsize
	\begin{tabular}{l|c|rrr|rrr|rrr}\toprule
		& &\multicolumn{3}{c|}{\shortstack{DONeRF~\cite{neff2021donerf} Dataset\\($800\times800$)}} &\multicolumn{3}{c|}{\shortstack{LLFF~\cite{mildenhall2019llff} Dataset\\($1008\times756$)}} &\multicolumn{3}{c}{\shortstack{LLFF~\cite{mildenhall2019llff} Dataset\\($504\times378$)}} \\\midrule
		Method &\shortstack{Memory\\{[MB]}} &\shortstack{Samples\\ per Ray} &\shortstack{Time\\{[ms]$\downarrow$}} &\shortstack{Quality\\PSNR$\uparrow$} &\shortstack{Samples\\ per Ray} &\shortstack{Time\\{[ms]$\downarrow$}} &\shortstack{Quality\\PSNR$\uparrow$} &\shortstack{Samples\\ per Ray} &\shortstack{Time\\{[ms]$\downarrow$}} &\shortstack{Quality\\PSNR$\uparrow$} \\\midrule
		\ours &\bron{4.1} &1.9 &\gold{31.3} &24.8 &2.0 &\gold{38.0} &22.0 &2.0 &\gold{9.9} &21.8 \\
        \ours &\bron{4.1} &3.7 &\bron{48.2} &27.5 &3.9 &\bron{58.9} &24.0 &3.9 &\silv{15.2} &23.3 \\
        \ours &\bron{4.1} &7.0 &78.9 &29.5 &6.9 &92.7 &25.2 &7.0 &\bron{24.4} &\bron{25.1} \\
        \ours &\bron{4.1} &12.6 &130.6 &\bron{30.8} &10.2 &129.6 &\bron{25.7} &10.6 &36.1 &\gold{26.2} \\\midrule
        \donerf &\bron{4.1} &2.0 &51.3 &27.9 &2.0 &61.1 &20.9 &- &- &- \\
        \donerf &\bron{4.1} &4.0 &86.3 &28.8 &4.0 &102.7 &21.6 &- &- &- \\
        \donerf &\bron{4.1} &8.0 &156.3 &29.8 &8.0 &186.1 &22.3 &- &- &- \\
        \donerf &\bron{4.1} &16.0 &296.2 &\silv{30.9} &16.0 &352.7 &22.9 &- &- &- \\\midrule
        \donerftermi &\bron{4.1} &2.0 &51.3 &27.2 &2.0 &61.1 &21.7 &- &- &- \\
        \donerftermi &\bron{4.1} &4.0 &86.3 &28.2 &4.0 &102.7 &22.3 &- &- &- \\
        \donerftermi &\bron{4.1} &8.0 &156.3 &29.2 &8.0 &186.1 &23.0 &- &- &- \\
        \donerftermi &\bron{4.1} &16.0 &296.2 &29.8 &16.0 &352.7 &23.6 &- &- &- \\\midrule
        NeRF &\silv{3.8} &256.0 &2360.7 &\silv{30.9} &256.0 &2810.9 &\gold{26.5} &- &- &- \\\midrule
        AutoInt &4.5 &- &- &- &- &- &- &16.0 &44.6 &24.1 \\
        AutoInt &4.5 &- &- &- &- &- &- &32.0 &88.5 &24.9 \\
        AutoInt &4.5 &- &- &- &- &- &- &64.0 &176.4 &\silv{25.5} \\\midrule
        Plenoxels &198.7 &- &\silv{47.9} &27.1 &- &\silv{51.3} &24.3 &- &- &- \\
        Plenoxels-MSI &892.9 &- &\silv{47.5} &29.6 &- &- &- &- &- &- \\
        Plenoxels-Large &3629.8 &- &- &- &- &110.1 &\silv{26.3} &- &- &- \\\midrule
        Instant-NGP-$2^{14}$ &\gold{2.0} &- &102.1 &29.4 &- &100.7 &24.8 &- &- &- \\
        Instant-NGP-$2^{19}$ &64.0 &- &161.8 &\gold{33.1} &- &137.0 &25.6 &- &- &- \\
        \bottomrule
	\end{tabular}
\end{table}

\newcommand\imgsize{0.19\linewidth}
\newcommand\tableimgsize{8.41em}

\newcommand{\addpic}[1]{\includegraphics[width=\tableimgsize]{#1}}

\newcolumntype{C}{>{\centering\arraybackslash}m{\tableimgsize}}
\begin{figure}
\centering
\captionsetup[subfigure]{labelformat=empty,justification=centering}
	\begin{minipage}{\linewidth}
	    \hfill
		\begin{subfigure}[t]{0.2555\linewidth}
		    \caption{\emph{Flower}}
			\includegraphics[width=\linewidth]{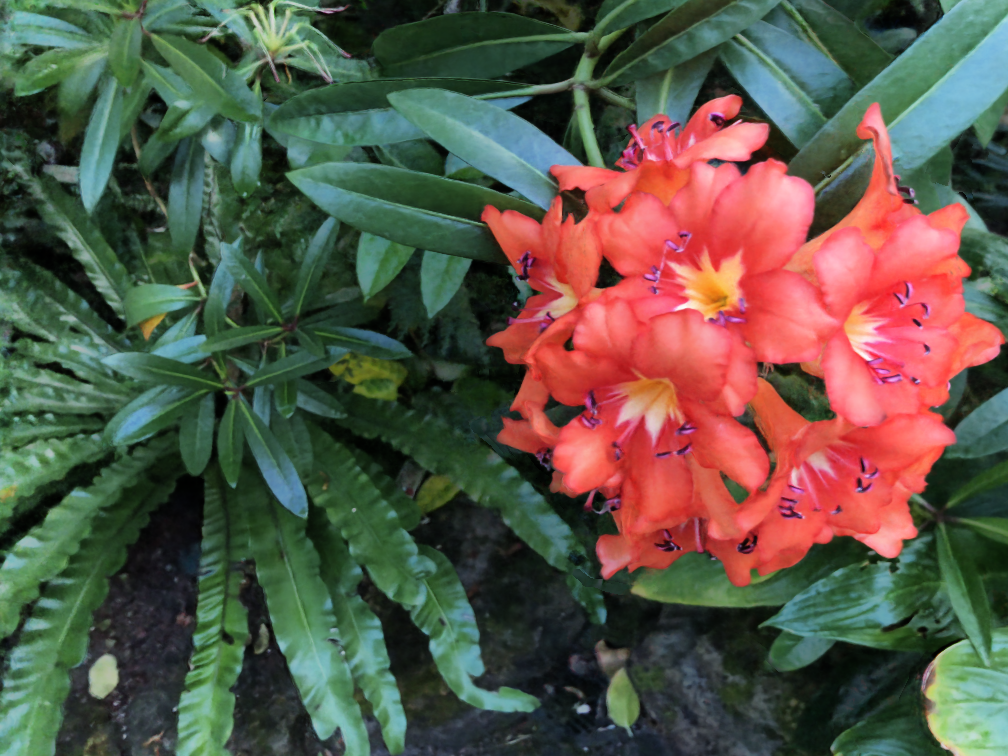}
		\end{subfigure}
		\begin{subfigure}[t]{0.2555\linewidth}
		    \caption{\emph{T-Rex}}
			\includegraphics[width=\linewidth]{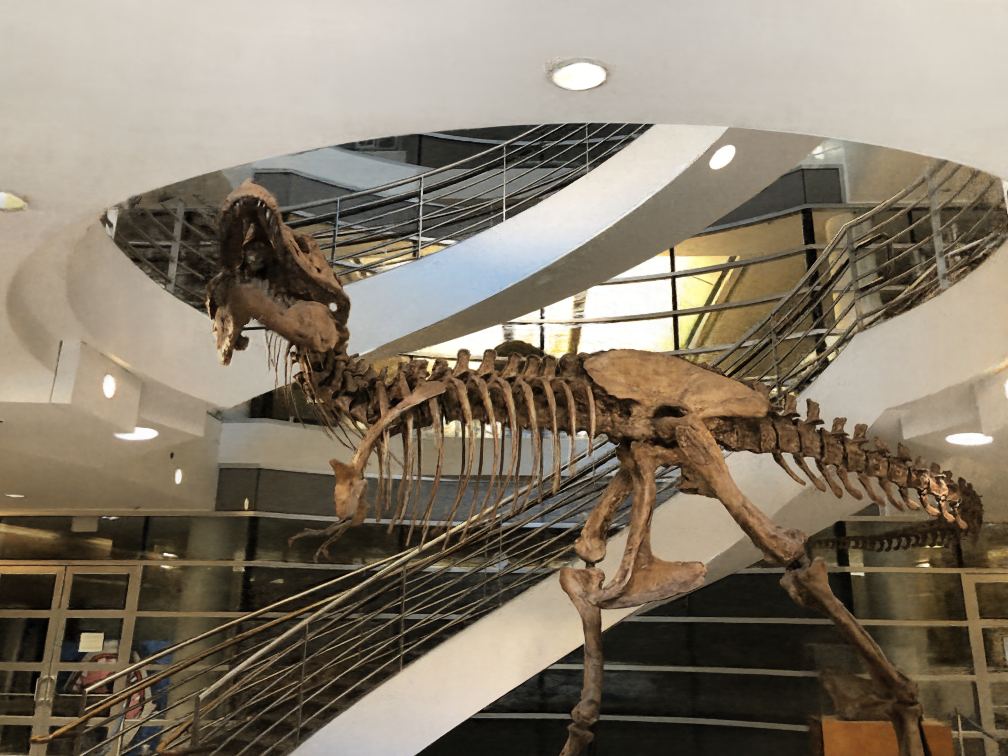}
		\end{subfigure}
		\begin{subfigure}[t]{0.1910\linewidth}
		    \caption{\emph{Pavillon}}
			\includegraphics[width=\linewidth]{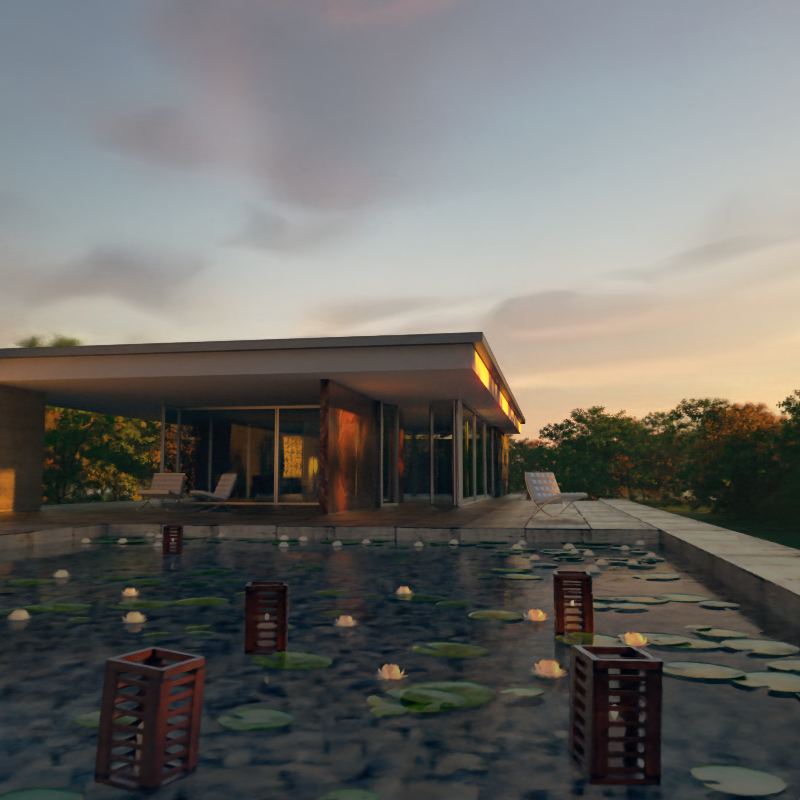}
		\end{subfigure}
		\begin{subfigure}[t]{0.1910\linewidth}
		    \caption{\emph{Classroom}}
			\includegraphics[width=\linewidth]{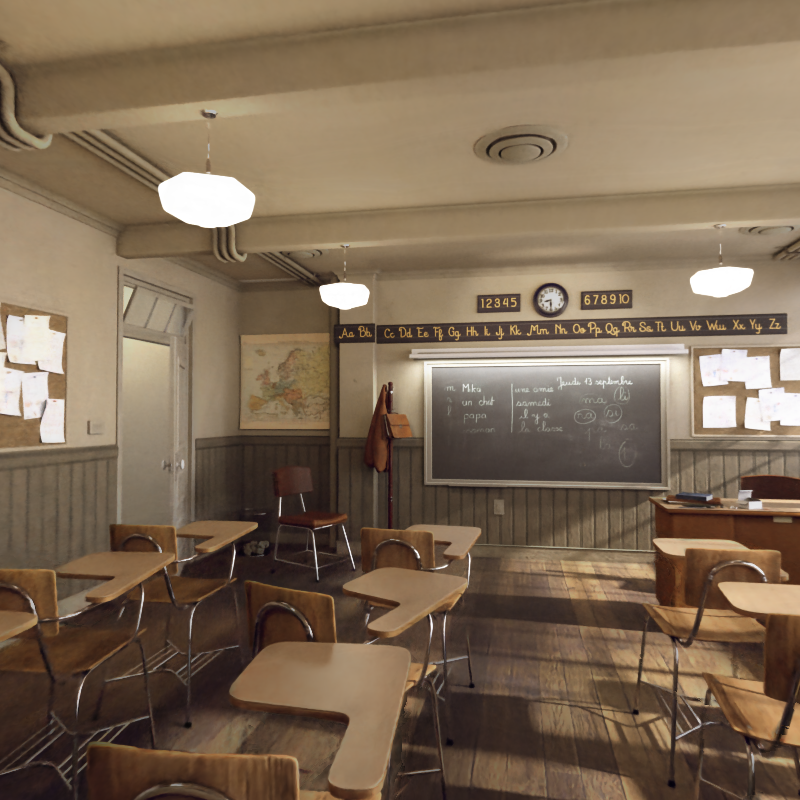}
		\end{subfigure}
		\begin{minipage}{0.000001\linewidth}
		\end{minipage}
	\end{minipage}\\
\begin{tabular}{r*4{C}@{}}
\rotatebox[origin=c]{90}{Ground Truth} & \addpic{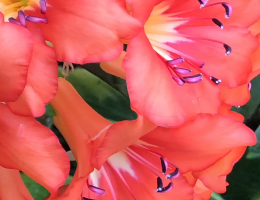} & \addpic{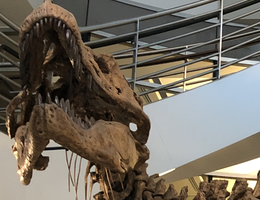} & \addpic{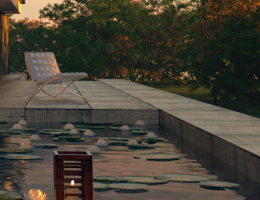} & \addpic{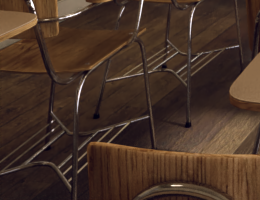} \\ 
\rotatebox[origin=c]{90}{\shortstack{\ours \\$\approx11$ samples}} & \addpic{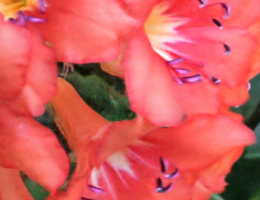} & \addpic{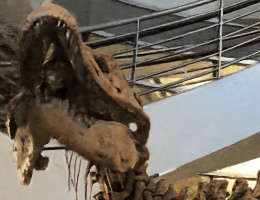} & \addpic{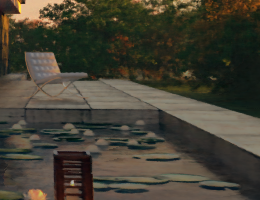} & \addpic{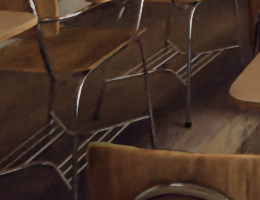} \\ 
\rotatebox[origin=c]{90}{\shortstack{\donerftermi \\ 16 samples}} & \addpic{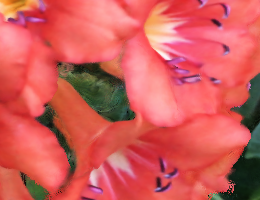} & \addpic{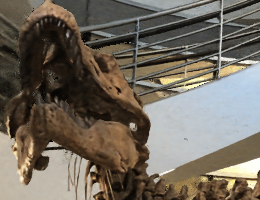} & \addpic{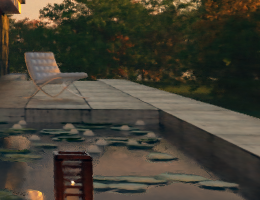} & \addpic{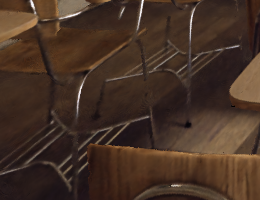} \\ 
\rotatebox[origin=c]{90}{\shortstack{NeRF \\ 256 samples}} & \addpic{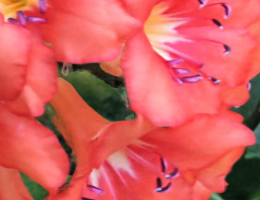} & \addpic{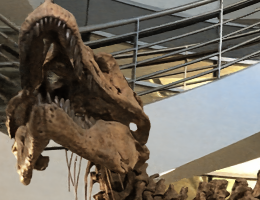} & \addpic{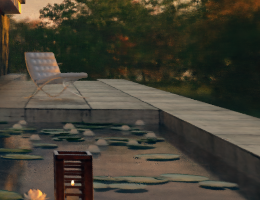} & \addpic{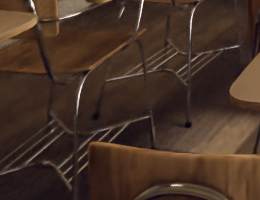} \\
\rotatebox[origin=c]{90}{\shortstack{Plenoxels\\ $\approx200$ MB}} & \addpic{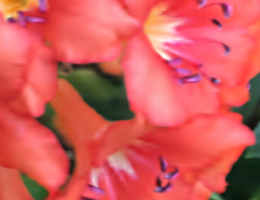} & \addpic{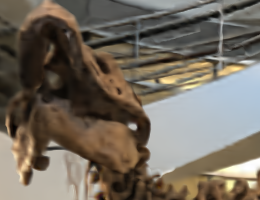} & \addpic{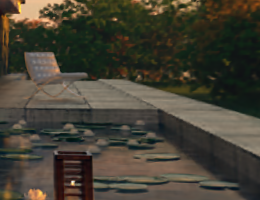} & \addpic{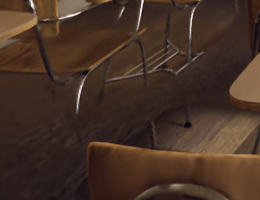} \\ 
\rotatebox[origin=c]{90}{\shortstack{Instant-NGP\\ Size $2^{19}$}} & \addpic{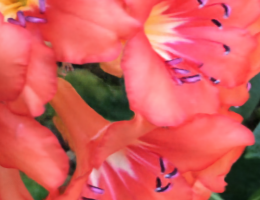} & \addpic{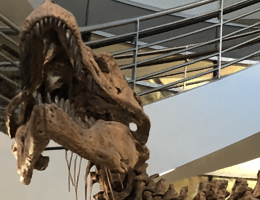} & \addpic{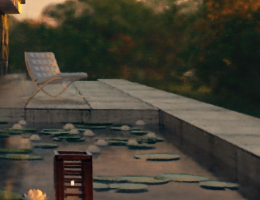} & \addpic{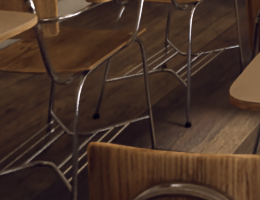} \\
\end{tabular}

\caption{\revised{Details on four test scenes, showing that \ours is similar in quality to the significantly slower NeRF and outperforms \donerftermi at lower sample counts.
While using $50\times$ more memory, Plenoxels tends to blur or leave out geometry due to lack of resolution in its grid. 
Instant-NGP achieves similar quality to \ours, while being slightly slower and requiring $16\times$ more memory.
}}
\label{fig:eval_qualitative}
\end{figure}

\subsubsection{Real-time Rendering Performance}
\label{sec:rendering_performance}
We evaluate the real-time rendering performance of our \ours TensorRT and CUDA prototype against all baselines, see Table~\ref{tbl:eval_results} (Columns ``Time'').
For all baselines except for Plenoxels, we evaluate their optimal rendering performance by computing the maximum throughput of identical networks in TensorRT, conservatively ignoring any additional input processing or differences in encoding.
For Plenoxels, we evaluate the rendering performance using the authors' provided implementation.

In terms of real-time rendering performance vs. memory footprint, \ours at a maximum sample count of $2$ achieves the best trade-off, being able to render scenes at an average frame rate of $26$ frames per second at a resolution of $1008\times756$.
The increased efficiency compared to \donerf and \donerftermi (at identical sample counts) comes from the optimized adaptive sampling kernels of \ours, which can lead to a massive speedup.
At equal rendering performance, \ours achieves significantly better quality compared to previous sampling-network based approaches, such as \donerf, \donerftermi and AutoInt on both the \donerf and LLFF datasets. 
At equal or improved quality, \ours outperforms \donerf by up to $5\times$, \donerftermi by up to \revised{$6\times$} and AutoInt by up to $7\times$.
Compared to the densely sampled NeRF, our largest \ours shows a $20\times$ increase in run-time, and the smallest \ours outperforms NeRF by up to $74\times$.
\revised{%
Compared to the highly optimized Instant-NGP, \ours achieves a comparable trade-off between quality and run-time, achieving up to a $2\times$ increase in run-time at equal quality.
}
The best trade-off between quality and run-time is reached by Plenoxels, which represent their scene within a large sparse grid, with an additional optional multi-sphere image background model. 
Although this enables real-time rendering for single scenes, the immense memory requirements of up to multiple gigabytes prevent use-cases such as streaming or splitting complex environments into multiple representations. 

The breakdown of frame time for the individual stages of \ours at a sample count of $2$ is given as: $1.54$ ms to generate the encoded inputs for the sampling network, $10.86$ ms to evaluate the sampling network, $1.02$ ms to generate the adaptively sampled inputs for the shading network, $18.16$ ms to compute the shading network inference, and $0.38$ ms for the final ray accumulation.
Thus, our adaptive sampling kernels and overall pipeline exhibit only minor overheads compared to the inference workload, which constitutes most of the frame time.
Furthermore, with higher average sample counts, the majority of additional compute load is added to the inference stages, and does not increase the overhead of input processing.
Overall, \ours fills a gap in the performance-quality-memory trifecta, being extremely fast at a compact memory footprint, at a low cost in image quality for certain scenes.

\section{Limitations and Future Work}
Although \ours already achieves promising results on real-world data, our evaluation does not optimize for camera parameters, and thus can suffer from input data that is not perfectly consistent.
Especially for very low sample counts (see Tab.~\ref{tbl:eval_results}), getting precise surface information is crucial to achieve good quality, and adding an additional optimization step for camera parameters and/or consistency in lighting could further improve results.
Second, the threshold for adaptive sampling, as well as the weights of the main optimization function (maximizing the sampling network's sparsity while preventing a collapse) influences the sparsity of \ours, affecting the overall quality and real-time performance.
While these parameters can be fairly robustly applied across different datasets, a grid search is recommended for best performance.
In the future, these parameters could be learned from data to save the time for hyperparameter tuning.

\section{Conclusion}
\label{sec:conclusion}

We have introduced \ours, a compact real-time dual-network neural representation that can be trained fully end-to-end via a soft student-teacher optimization scheme.
It is the first of its kind to adaptively place a very low amount of samples for each individual ray.
We significantly outperform previous work that utilized sampling networks for very low sample count neural representations. %
Due to the compact nature of our neural representation, we additionally showed how multiple models can be blended in overlapping regions, which opens the door for real-time rendering of dynamically streamed neural representations of complex environments.
We believe that such a fully neural real-time representation can be a useful alternative to approaches that require explicit spatial data structures.

\bibliographystyle{splncs04}
\bibliography{6513.bib}

\end{document}